\documentclass[letterpaper, 10 pt, conference]{ieeeconf}  %
\pdfoutput=1

\IEEEoverridecommandlockouts                              

\usepackage[backend=bibtex8,
            hyperref=false,
            url=false,
            isbn=false,
            doi=false,
            backref=false,
            style=ieee,
            citestyle=numeric-comp,
            sorting=nyt,
            block=none]{biblatex}

\renewcommand{\bibfont}{\small}
\usepackage{graphics}
\usepackage[pdftex]{graphicx}
\usepackage{wrapfig}
\DeclareGraphicsExtensions{.pdf,.png,.jpg}
\usepackage{epsfig}
\usepackage[font={small}]{caption}
\usepackage{subcaption}
\usepackage[rightcaption]{sidecap}
\usepackage{pbox}

\usepackage{bigstrut}
\usepackage{mathtools}
\usepackage{amsmath, amssymb, amscd}
\usepackage{ wasysym } 
\usepackage{amsfonts}
\usepackage{mathptmx} 
\usepackage{gensymb} 
\usepackage{nicefrac}       
\numberwithin{equation}{section} 

\DeclareMathAlphabet{\mathcal}{OMS}{lmsy}{m}{n}
\DeclareSymbolFont{largesymbols}{OMX}{cmex}{m}{n}
\usepackage{textcomp} 

\usepackage{algorithm} 
\usepackage{algorithmicx, algpseudocode} 

\usepackage{array} 
\usepackage{tabularx}
\usepackage{multirow}
\usepackage{multicol}
\usepackage{booktabs}
\usepackage{tabulary}

\usepackage[T1]{fontenc} 
    \usepackage{ae} %
    \usepackage{aecompl}
\usepackage[utf8]{inputenc}
\usepackage[protrusion=true,expansion=true]{microtype}
\usepackage[english]{babel} 
\usepackage{units}
\usepackage{siunitx}
\usepackage{bm}
\usepackage{times} 
\usepackage{xspace}
\usepackage{flushend}
\usepackage{balance} 
\usepackage{csquotes}
\usepackage{makeidx}
\usepackage{blindtext}

\let\labelindent\relax
\usepackage{enumitem}

\usepackage{tikz}
\usetikzlibrary{backgrounds}

\usepackage{ragged2e}
\usepackage{soul} 
\usepackage{subfiles} 

\usepackage[yyyymmdd]{datetime}

\date{\protect\formatdate{1}{1}{2001}}

\usepackage{url}
\makeatletter
\g@addto@macro{\UrlBreaks}{\UrlOrds}
\makeatother
\usepackage{color}
\usepackage{pgfplots}
\pgfplotsset{compat=newest}

\usepackage{hyperref}
\hypersetup{
    colorlinks=true,
    linkcolor=black,
    citecolor=black,
    filecolor=cyan,
    urlcolor=black
}

\usepackage{marginnote}
\usepackage{soul} 

\usepackage[colorinlistoftodos,prependcaption,textsize=small]{todonotes}
\usepackage{regexpatch}
\makeatletter
\xpatchcmd{\@todo}{\setkeys{todonotes}{#1}}{\setkeys{todonotes}{inline,#1}}{}{}
\makeatother

\newcommand{\tocite}[1]{%
\textcolor{red}{[cite:\ifthenelse{\equal{#1}{}}{}{#1}?]}
}

\newcommand{\ignore}[1]{}
\newcommand{\etal}{et al.~}

\renewcommand{\baselinestretch}{0.9122}



\DeclareMathOperator*{\argmax}{\arg\!\max}

\newcommand{\underscoredw}[1]{\textrm{#1}}

\newcommand{\policya}{{a}}

\newcommand{\controlleraB}{{u}}

\newcommand{\qdes}{{q}_{\underscoredw{des}}} 
\newcommand{\qdot}{{\dot q}} 
\newcommand{\qdotdes}{{\dot q}_{\underscoredw{des}}}
\newcommand{\taudes}{{\tau}_{\underscoredw{des}}}
\newcommand{\q}{{q}} 
\newcommand{\qdelta}{\Delta{q}} 

\newcommand{\massmatrix}{{M}}
\newcommand{\lambdapos}{{\Lambda}^{\underscoredw{pos}}}
\newcommand{\lambdaori}{{\Lambda}^{\underscoredw{ori}}}
\newcommand{\jacpos}{{J}^{T}_{\underscoredw{pos}}}
\newcommand{\jacori}{{J}^{T}_{\underscoredw{ori}}}

\newcommand{\oridelta}{\Delta^{\underscoredw{ori} }}
\newcommand{\ori}{{R}}

\newcommand{\orides}{{R}_{\underscoredw{des}}}
\newcommand{\orivel}{{\omega}}
\newcommand{\kppos}{{k}_{\underscoredw{p}}^{\underscoredw{pos}}}
\newcommand{\kvpos}{{k}_{\underscoredw{v}}^{\underscoredw{pos}}}
\newcommand{\kpori}{{k}_{\underscoredw{p}}^{\underscoredw{ori}}}
\newcommand{\kvori}{{k}_{\underscoredw{v}}^{\underscoredw{ori}}}

\newcommand{\cartdelta}{\Delta^{\underscoredw{pos} }}

\newcommand{\cartposd}{\mathbf{x}_{\underscoredw{des}}}

\newcommand{\cartvel}{{v}}

\newcommand{\posdes}{{p}_{\underscoredw{des}}}
\newcommand{\pos}{{p}}

\newcommand{\kp}{{k}_{\underscoredw{p}}}
\newcommand{\kv}{{k}_{\underscoredw{v}}}

\newcommand{\taurobot}{{u}}

\renewcommand{\theequation}{\arabic{equation}}
\title{\LARGE \bf
Variable Impedance Control in End-Effector Space:\\An Action Space for Reinforcement Learning in Contact-Rich Tasks}

\author{Roberto Mart\'in-Mart\'in, Michelle A. Lee, Rachel Gardner, Silvio Savarese, Jeannette Bohg, Animesh Garg
\thanks{All authors are with Stanford Artificial Intelligence Lab (SAIL), Stanford University. A. Garg is also with Nvidia, USA. 
This work has been partially supported by JD.com American Technologies Corporation (``JD'') under the SAIL-JD AI Research Initiative. This article solely reflects the opinions and conclusions of its authors and not JD or any entity associated with JD.com.
{\tt\small [robertom, mishlee, rachel0, ssilvio, bohg, animeshg]@stanford.edu}
}%
}

\begin{document}
\maketitle
\thispagestyle{empty}
\pagestyle{empty}

\begin{abstract}
Reinforcement Learning (RL) of contact-rich manipulation tasks has yielded impressive results in recent years. While many studies in RL focus on varying the observation space or reward model, few efforts focused on the choice of action space (e.g. joint or end-effector space, position, velocity, etc.). 
However, studies in robot motion control indicate that choosing an action space that conforms to the characteristics of the task can simplify exploration and improve robustness to disturbances.
This paper studies the effect of different action spaces in deep RL and advocates for \ul{v}ariable \ul{i}mpedance \ul{c}ontrol in \ul{e}nd-effector \ul{s}pace (VICES) as an advantageous action space for constrained and contact-rich tasks.
We evaluate multiple action spaces on three prototypical manipulation tasks: Path Following (task with no contact), Door Opening (task with kinematic constraints), and Surface Wiping (task with continuous contact). We show that VICES improves sample efficiency, maintains low energy consumption, and ensures safety across all three experimental setups. Further, RL policies learned with VICES can transfer across different robot models in simulation, and from simulation to real for the same robot. Further information is available at \href{https://stanfordvl.github.io/vices}{https://stanfordvl.github.io/vices}.
\end{abstract}

\section{Introduction}
\label{sec:intro}

Diverse control tasks in robot manipulation are naturally addressed in different action spaces -- for a specific task, one action space might simplify learning and control more than another.
For a walking robot, it is important to directly control contact interactions to avoid slippage~\cite{Ludo_Contact_2013}. 
In contrast, for a tennis swing, it is important to track and control position, velocity, and at times the acceleration of the end-effector~\cite{DMP_Initial_2002}. 
For a surface-to-surface alignment task, minimizing the moment around a contact is important for robustness~\cite{khansari2016adaptive}.
Tackling all these tasks requires solving two subproblems: (1) the generation of reference signals (desired contacts, trajectories, moments, etc.), and (2) the tracking of these signals.

Control systems for robots can be structured with two feedback control loops that address the aforementioned subproblems: an outer loop controller that generates a time-varying reference trajectory, and an inner loop controller that tracks this trajectory. Let us refer to the outer loop as a functional map from observations to reference signals $g(o): \mathcal{O} \rightarrow \mathcal{A}$, and the inner loop as a map from reference signals to actuation commands $f(a): \mathcal{A} \rightarrow \mathcal{U}$. The combined control law becomes: $u = f\circ g (o)$, where $o \in \mathcal{O}$ is some observation, $a \in \mathcal{A}$ is an abstract action providing a reference signal in some space, and $u \in \mathcal{U}$ is the control command sent to the robot's actuators to track this reference. 

While control theory provides a vast repertoire of strategies to map from reference signals to actuation commands ($f(\cdot)$ implementations), a key problem in robotics is to generate a viable reference in a suitable space given raw sensory observations, i.e. modelling the function $g(\cdot)$ and deciding on the interface to $f(\cdot)$. Once the interface has been defined (e.g. forces, positions, contact points, etc.), the generation of reference signals by $g(\cdot)$ can be addressed in multiple ways, ranging from hand-tuned state machines~\cite{sen2016automating,khansari2016adaptive}, trajectory optimization~\cite{18-toussaint-RSS,2017_rss_system}, imitation learning~\cite{schaal2003computational,Ijspeert:2013:DMP,Calinon08IROS,kroemer2015towards}, or reinforcement learning (RL)~\cite{lee2019making,harrison2017adapt}. 

Recent research in RL has focused on ``observations-to-torques''~\cite{levine2016end}, which is akin to merging $f(\cdot)$ and $g(\cdot)$ into a single learned model. 
Other methods use higher-level action spaces, such as joint space commands (e.g. position or velocity)~\cite{gu2017deep,haarnoja2018soft,vevcerik2017leveraging,zhu2018reinforcement} or task space commands (e.g. end-effector poses, force or fixed impedance)~\cite{lee2019making,Mrinal:2011}. These works typically focus on the effects of choosing an observation space $\mathcal{O}$ on the learning process and rarely justify the choice of action spaces, $\mathcal{A}$. 
However, the choice of action space defines the quantity around which the inner control loop is closed, and by extension the space wherein tracking error is minimized. This critically impacts robustness and task performance as well as learning efficiency and exploration. 

\begin{figure}[t]
\centering
\includegraphics[width=0.43\textwidth]{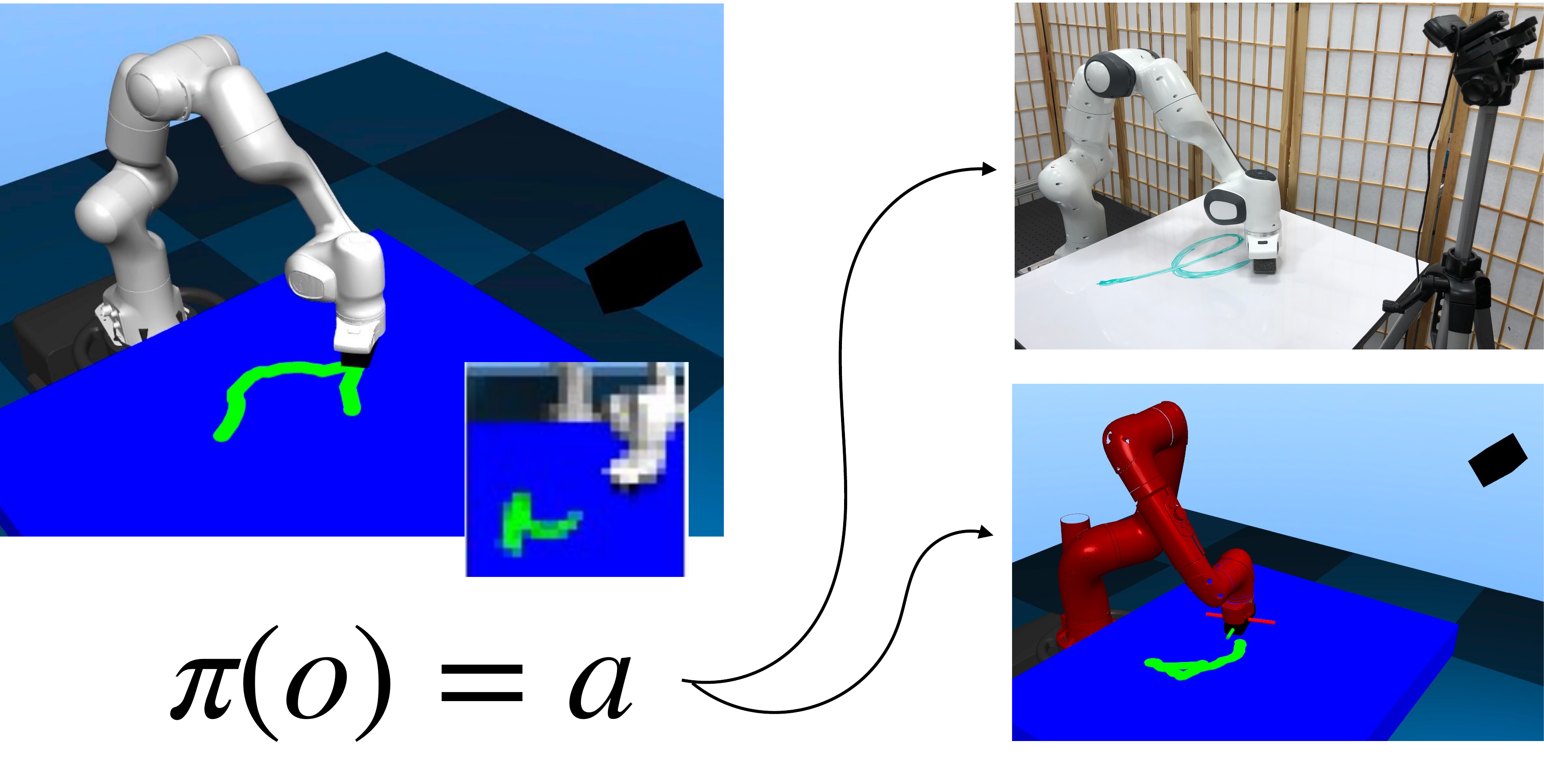}
\caption{Depending on the characteristics of a task, different action spaces for policy learning are better suited than others; \textit{Variable impedance control in end-effector space (VICES)} is efficient to learn both free-space motion and contact-rich tasks. Through the compensation of the robot's kinematics and dynamics, the policies learned for one robot in simulation transfer seamlessly to other robot instances and to the real robot.}
\label{fig1}
\end{figure}

Moreover, previously proposed action spaces do not necessarily create suitable references for some contact-rich tasks. Consider the task of wiping a board, wherein a robot must control forces in some directions (to keep pressing against the board) and motion in others. 
The physical constraints of the task dictate in which axes the robot should be stiff and in which it should be compliant. This can often be time-varying. Manual specification of the task constraints is not a scalable solution for the variety of contact-rich tasks the robot may need to perform, and for some tasks, manual specification may be non-trivial. 



This paper studies how the selection of an interface between $g(\cdot)$ and $f(\cdot)$, (the space $\mathcal{A}$) affects RL as a method to learn the mapping from observations to reference signals, $g(\cdot)$,
and presents the first empirical study comparing the most common choices for the $\mathcal{A}$ in contact-rich manipulation. 
We argue that an action space that captures motion and impedance in end-effector space can enable efficient learning of such tasks. 
We evaluate joint position, joint velocity, joint torque, joint variable impedance, as well as fixed and variable impedance in end-effector space.
The choice of action space should be guided not only by the robot model but also by prior knowledge of the task. Hence, we compare action spaces across tasks with varying degrees of task-space constraints, i.e., Path Following with no contact (\textit{path following}), manipulation of constrained mechanisms (\textit{door opening}), and continuous unconstrained contact (\textit{surface wiping}).


Moreover, we introduce \textit{\ul{v}ariable \ul{i}mpedance \ul{c}ontrol in \ul{e}nd-effector \ul{s}pace (VICES)} 
and advocate this action space for Deep RL algorithms applied to contact-rich manipulation. 
We show that policies defined in VICES improve sample efficiency for exploration in RL, energy efficiency, and reduce applied forces. Thanks to the classical dynamically consistent operational space formalism~\cite{khatib1987unified}, we observe that policies learned in end-effector space are also more robust to transfer across robots with significant differences in dynamics whether in simulation or the real world. 


\section{Related Work}
\label{sec:tw}

\noindent \textbf{Robot Motion Control:}
Compliant control of a robot manipulator enables adaption to uncertainty in the environment (e.g. exact shape of a surface, or kinematic constraints of mechanisms) during contact-rich tasks. However, certain tasks require direct control of the contact interactions (e.g. don't apply too much force when wiping a window). Previous abstractions have divided dimensions of the task into those that are controlled kinematically (through position and velocity) and dynamically (through force and torques)~\cite{4308708,khatib1987unified,kroger2004compliant}. However in practice, the hard decoupling requires a level of knowledge about the task that is not always available~\cite{508440}. Impedance control~\cite{part1985impedance} allows safe robot contact manipulation with an unknown environment by explicitly controlling the amount of force the robot exerts when it deviates from a given kinematic goal. Therefore, it alleviates the need for perfect knowledge or hard separation between the dynamic and kinematic task dimensions. 

However, different phases of a manipulation task may require a dynamic balancing between kinematic and dynamic control. Existing methods address it by scheduling variable impedance gains to maintain stability or safety for a given kinematic trajectory~\cite{Mitrovic2011,Li2018,ruckert2013learned}. However, these methods assume that a reference trajectory is given.
Instead, we propose to directly predict both end-effector displacement (reference) and variable impedance gains based on observations. 



\vspace{2pt}
\noindent \textbf{Action Spaces in Learning from Demonstrations:}
LfD derives a task policy based on demonstrations provided by some other agent(s)~\cite{Argall:2009:SRL}. If the demonstrations do not perfectly overlap, a possible approach is to derive a policy that imitates the mean motion of the demonstration set, and varies the stiffness according to the coherence of the trials~\cite{6636303,5648931} or according to the force sensed during kinesthetic replay~\cite{Abu-Dakka2018}. Similar use of variable impedance as an action representation for LfD has been demonstrated to be successful for adaptive grasping~\cite{6907861}, manipulation of deformable objects~\cite{lee2015learning}, and co-adaptation to human workers~\cite{rozo2016learning}. 
However, the specification of impedance in the demonstration only reflects variability in demonstrations trajectories but not the underlying task constraints imposed by the environment nor the force profiles required for the task. 
This approach is also restricted to tasks where expert demonstrations are feasible, and hence is limited in application to kinematic tasks with phases that require different level of precision.  


\vspace{2pt}
\noindent \textbf{Reinforcement Learning:}
In the field of model-based reinforcement learning, Kim \etal\cite{kim2010impedance} proposed a method to learn the parameters of a variable impedance position controller in end-effector space based on the equilibrium point formalism. They demonstrated the convergence, robustness and energy efficiency of their method on simulated manipulation task with a two DoF planar arm. However, their method requires an initial trajectory which is not always available.

Similar to our method, Buchli \etal\cite{buchli2011learning} apply policy improvements with path integrals (PI$^2$)~\cite{theodorou2010generalized} to refine initial trajectories and learn variable scheduling for the joint impedance parameters. They demonstrate that energy consumption can be optimized while achieving a task using variable impedance. However, they use joint space as their action space, which limits the transferability of the learned behaviors to different robots and the optimality of the trajectories in the space of the task. Also, their method requires an initial estimate of the solution to start the iteration. 

Rey \etal\cite{Rey2018} propose an approach to simultaneously learn kinematic trajectories from demonstrations and variable impedance in task space from exploration. They use Gaussian Mixture Regression as representation for the policy and demonstrate their method in simulation and in one real-world planar task with one single stiffness parameter. \cite{Mrinal:2011} proposed a method to refine given trajectories with additional force profiles using PI$^2$\cite{theodorou2010generalized} and readings from a force-torque sensor. We aim to achieve dynamic behavior without direct force loop control by using impedance to learn both trajectories and variable stiffness profiles.

It is worth noting that all previous approaches have boostrapped learning with initial demonstrations while we explore learning from scratch to better understand how fast policies converge.
Viereck~\etal\cite{Viereck2018} studied how to incorporate control structure to learn hopping policies for one-legged robot with RL. They use an optimal controller for fixed task conditions and learn to imitate its policy with neural networks to generalize to new task conditions. Interestingly, they compare two network architectures outputting signals in different action spaces: directly desired torques or full feedback parameters and desired configuration, which is transformed into torques with an analytic function. Their experiments show that this second action space is best suited for the hopping task with intermittent contact and adds interpretability to the network output. We study a set of analytic functions (controllers) that map policy actions to low level robot commands for robot manipulation in three tasks with different contact properties.

With related motivation to ours, Peng \etal\cite{Peng:2017:LLS:3099564.3099567} studied the importance of different action representations in RL for the task of locomotion. Similar to us, they aimed to shed light on the best action space to be used, but in their case they focused on imitation learning in bipedal motion of simulation agents. We would like to provide similar insights in the more complex contact-rich robot manipulation domain and include preliminary studies of transfer to real world. 



\section{Reinforcement Learning}
\label{sec:rl}

The goal in reinforcement learning is to find a policy $\pi$, that selects actions based on current observations so as to maximize the expected reward obtained from interactions with the environment~\cite{sutton2018reinforcement}.  
We assume that the underlying problem can be modelled as a discrete-time continuous Markov decision process ($\mathcal{S}$, $\mathcal{A}$, $\mathcal{P}$, $\mathcal{R}$, $\gamma$, $\rho$), where $\mathcal{S}$ is a continuous state space, $\mathcal{A}$ is a continuous action space, $\mathcal{P}$ is a Markovian transition model defining the probability of transitioning between states for a given action $P(s'|s,a)$,  $\mathcal{R}$ is a reward function $\mathcal{R}(s,a) = r \in \mathbb{R}$, $\gamma \in [0,1)$ is a discount factor (for infinite horizon problems) and $\rho$ is the initial state distribution. When $\pi$ is probabilistic it represents the probability of the action, $a$, given the state, $s$:  $\pi(a|s) = p(a|s)$, and $\pi(\cdot|s)$ is the density distribution over $\mathcal{A}$ in state $s$. 

Alternatively, we can assume the state is not directly observable and learn a policy $\pi(a|o)$  conditioned on observations $o \in \mathcal{O}$ instead of latent states $s \in \mathcal{S}$.
Herein, the agent following the policy $\pi$ obtains an observation $o_t$ at time $t$ and performs an action $a_t$, receiving from the environment an immediate reward $r_t$ and a new observation $o_{t+1}$. Assuming the policy is parameterized by $\theta$, a policy gradient algorithm optimizes $\theta$ to maximize that the expected future return: 
\begin{equation}
\theta^* = \argmax_{\theta}\,J(\theta,\rho)= \argmax_{\theta}\, E\left[\sum_{t}\gamma^t r_t\right]
\end{equation}

\noindent These algorithms are based on the policy gradient theorem that states: $\frac{\delta J}{\delta\theta}=\sum_{s}\mu_{\pi_{\theta}}(s,\rho)\sum_{a}\frac{\delta\pi_{\theta}(a|s)}{\delta\theta}Q_{\pi_{\theta}}(s,a)$, where $\mu_{\pi_{\theta}}(s,\rho)=\sum_{t}\gamma^{t}P(s_t=s|\rho)$ and $Q_{\pi_{\theta}}$ is the action-value function associated to the current policy ${\pi_{\theta}}$.

There are multiple algorithmic solutions based on the policy gradient theorem that allow us to represent the policy with a deep neural network, e.g. Trust Region Policy Optimization (TRPO)~\cite{schulman2015trust}, Deep Deterministic Policy Gradients (DDPG)~\cite{lillicrap2015continuous}, or Advantage Actor-Critic (A2C)~\cite{mnih2016asynchronous}. In our evaluation of different action spaces for policies, we will use Proximal Policy Optimization (PPO)~\cite{schulman2017proximal}. The evaluation of the sensitivity of different algorithms to the action space is deferred to future work.

\section{Action Spaces in RL for Robot Manipulation}
\label{sec:method}
Relating the formalisms we introduced in Sec.~\ref{sec:intro} and~\ref{sec:rl}, $\pi$ corresponds to $g(\cdot)$, the function that maps observations to reference actions in some space $\mathcal{A}$, assuming that another function $f(\cdot)$ will map these actions to low level control commands, $u \in \mathcal{U}$.

We note that the RL algorithms in Sec.~\ref{sec:rl} are agnostic to choice of action space $\mathcal{A}$.
In practice, the most common $\mathcal{A}$ in RL for robot manipulation are a) joint torques~\cite{levine2016end}, b) joint velocities~\cite{gu2017deep,vevcerik2017leveraging,zhu2018reinforcement}, c) joint positions~\cite{haarnoja2018soft}, and d) end-effector position~\cite{lee2019making,thananjeyan2017multilateral} possibly with orientation.
The most common lowest level control commands (and the one we assume for our underlying physical agent) are joint torques, $u=\tau \in \mathcal{T}$. 
Joint torques are safer than positions and velocities for contact-rich tasks in unstructured environments, because the forces the robot will apply on the environment are limited by the specified desired torques. 

\begin{figure}[t!]
\centering
\includegraphics[width=0.92\linewidth,clip]{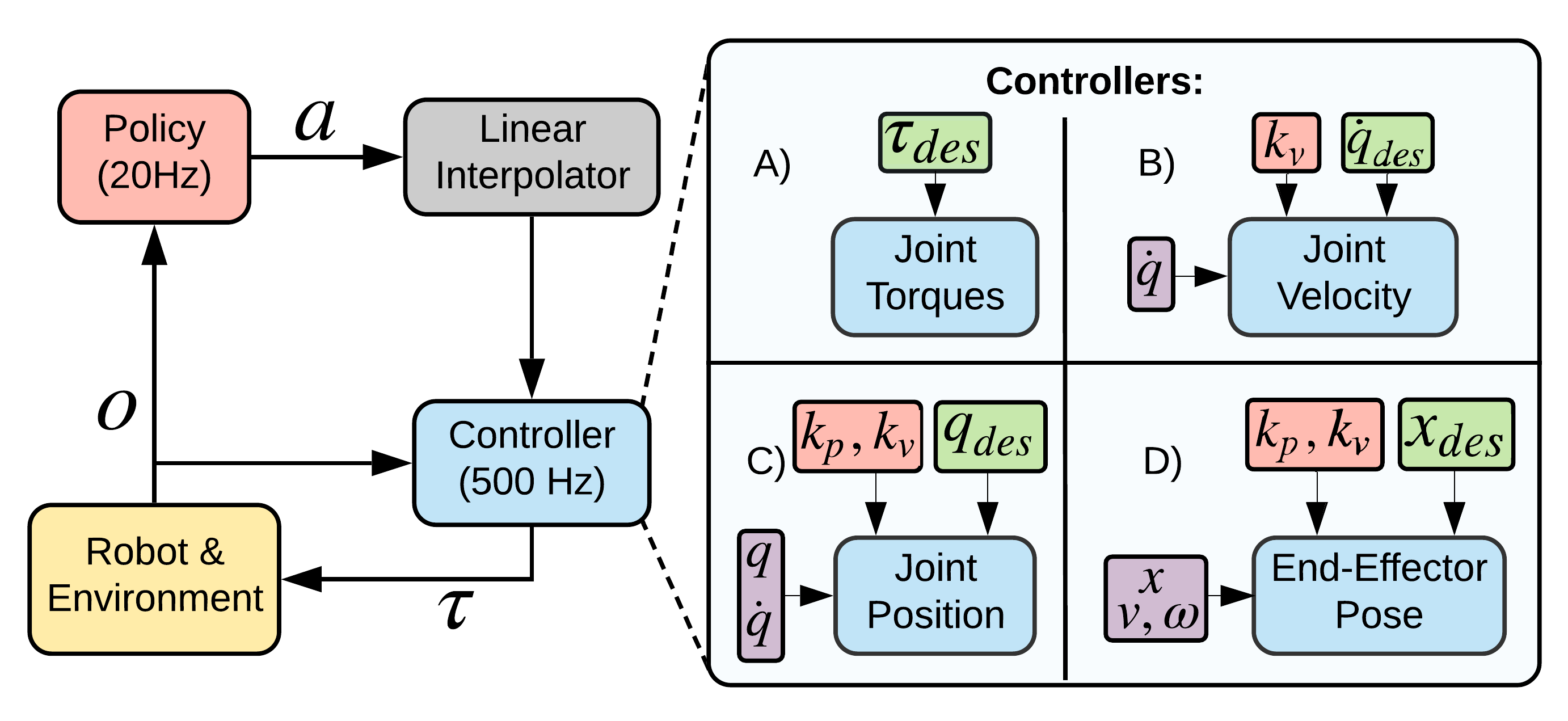}
\caption{We trained policies that output actions, $a$,  on different action spaces, $\mathcal{A}$ at \SI{20}{\hertz}; Depending on the action space, the appropriate controller takes in the desired control signal, the control parameters (fixed or given by the policy), and the current state and outputs actuation commands $\controlleraB\in\mathcal{U}$ at \SI{500}{\hertz}, which for our robots correspond to joint torques $\tau$. We interpolate linearly between consecutive low-frequency policy actions to generate smoother high-frequency controller signals (and parameters, if controlled by policy).}
\label{fig:controller}
\end{figure}

Manipulation tasks can seldom be solved solely by only controlling motion since there are tasks that contain contact and force constraints (e.g. the adaptation required to manipulate an articulated object or the minimum force to press while we wipe a surface)~\cite{4308708,khatib1987unified,508440,kroger2004compliant}. To succeed in these tasks, the robot needs to dynamically modulate the exerted force on the environment through the torques on each joint. 



To map between action space $\mathcal{A}$ and actuation space $\mathcal{U}$, we can define analytic parameterized functions (i.e. \emph{controllers}), $f_{\kappa}(\policya, s_{\textrm{\it robot}}): \mathcal{A} \times \mathcal{S}_{\textrm{\it{robot}}} \to \mathcal{U}$, that transform the output of the policy from the action space to the space of control commands depending on the current state of the robot, $s_{\textrm{\it robot}} \in \mathcal{S}_{\textrm{\it{robot}}}$. The parameters of these functions, $\kappa$, can be made part of the policy action space so that the agent has full controllability on the manipulation behavior~\cite{buchli2011learning}. 



In the following, we will introduce the different choices of analytic controllers $f_{\kappa}$, we use to map policy actions from commonly used policy action spaces into joint torques. 

\textbf{Joint Torques:}
When the policy directly outputs desired joint torques, i.e. $\policya=\taudes$, the function that transforms to robot commands is simply ($\textrm{jt} = \textrm{\it joint torques}$): 
\begin{equation}
    \taurobot = f^{\textrm{{jt}}}_{\kappa}(\policya=\taudes) = \taudes
\end{equation}

\begin{figure*}[ht!]
\begin{subfigure}[t]{0.32\textwidth}
\includegraphics[width=0.95\linewidth]{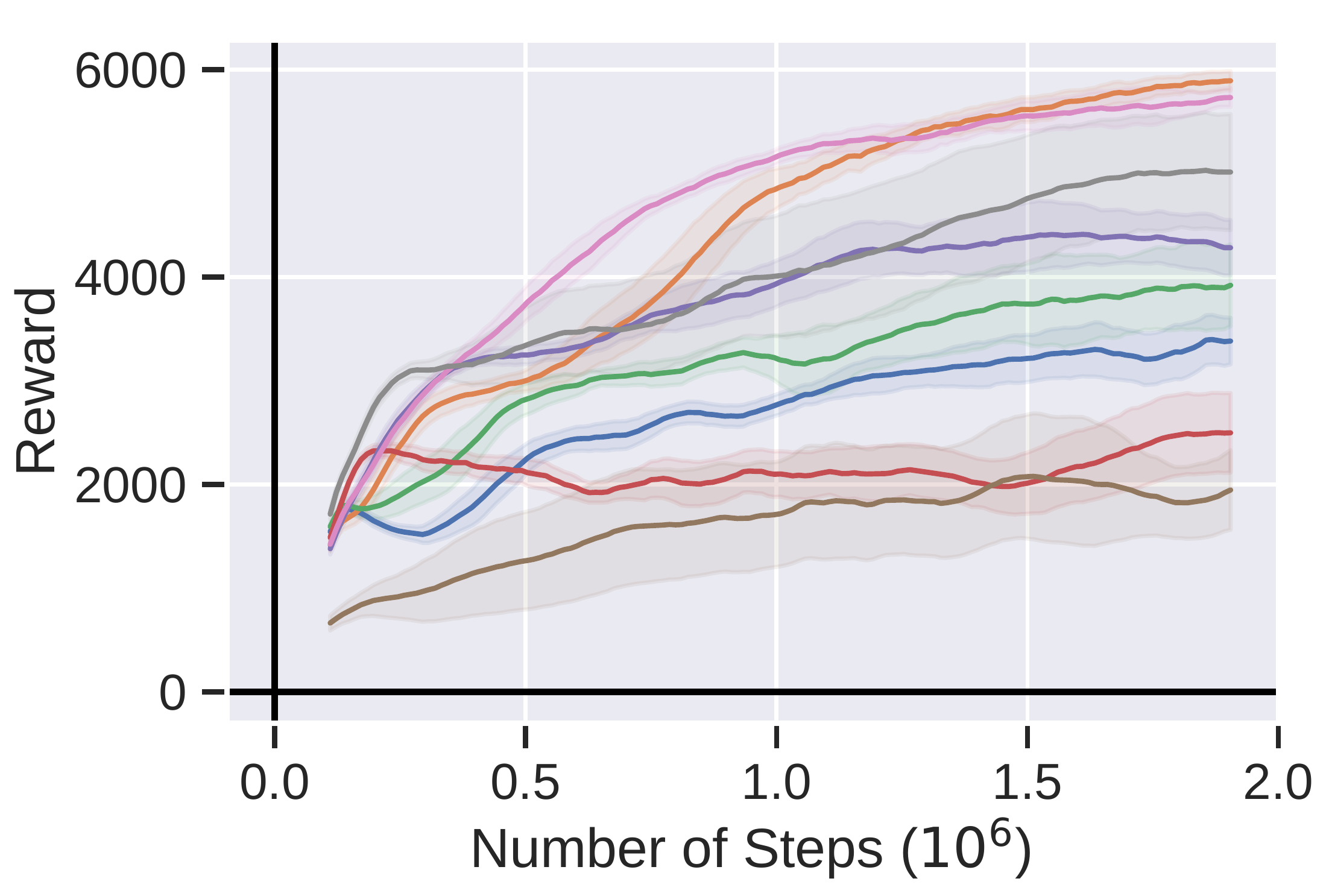}%
\caption{Path Following}
\end{subfigure}
\hfill
\begin{subfigure}[t]{0.32\textwidth}
\includegraphics[width=0.95\linewidth]{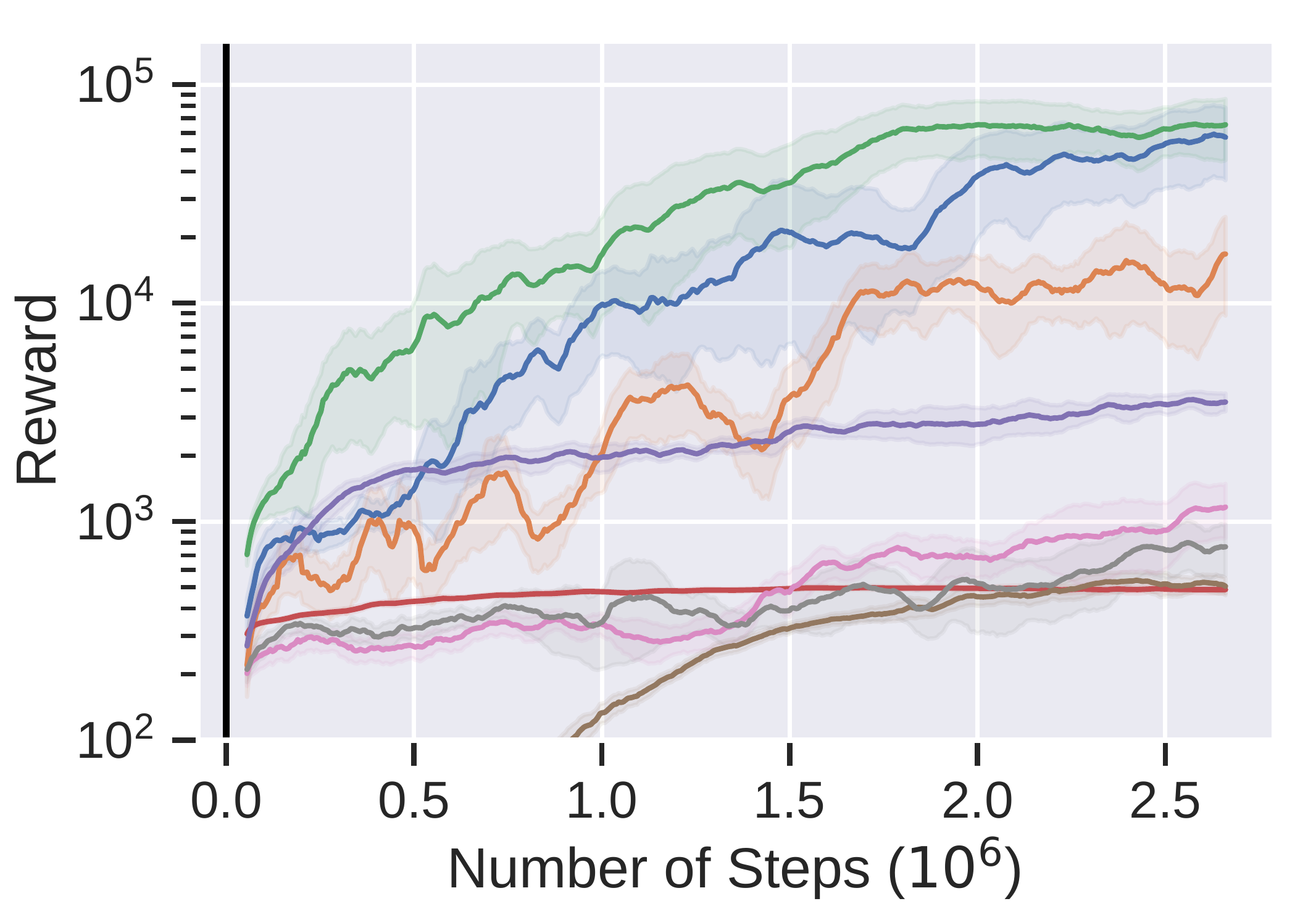}%
\caption{Door Opening}
\end{subfigure}
\hfill
\begin{subfigure}[t]{0.32\textwidth}
\includegraphics[width=0.95\linewidth]{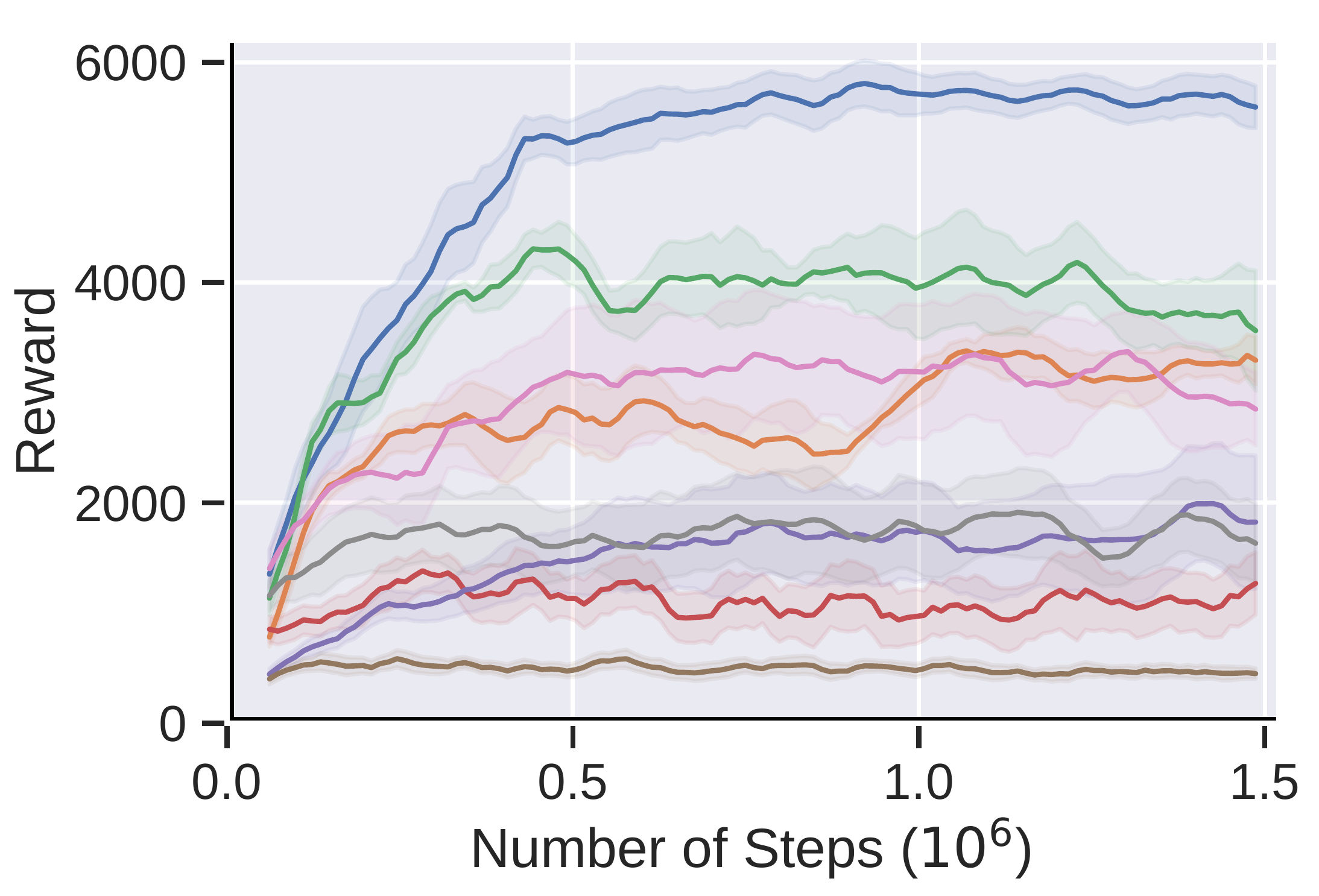}%
\caption{Surface Wiping}
\end{subfigure}
\\  
\vspace{-5pt}
\begin{subfigure}[t]{0.99\textwidth}
\vspace{-2pt}
\includegraphics[trim={0 5cm 0 0.5cm},clip,width=\linewidth]{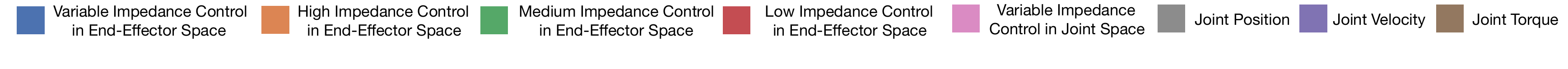}%
\end{subfigure}
\caption{Training curves for a) Path Following (free space), b) door opening (kinematic constraints), and c) surface wiping (contact rich) tasks; The plots depict mean and standard deviation of five learning processes with different random seeds; Tasks without contact or with kinematic constraints (Path Following and door opening) do not require variable impedance as action space to achieve high reward; In the contact-rich task (surface wiping) the policy using variable impedance in end-effector space achieves higher reward because it learns to adapt correctly the amount of force applied to the tasks constraints}
\vspace{-10pt}
\label{fig:training}
\end{figure*}

\textbf{Joint Velocities:}
When the policy outputs reference joint velocities $\policya = \qdotdes$, the function to map to joint torques is ($\textrm{jv} = \textrm{\it joint velocities}$):
\begin{equation}
\taurobot = f^{\textrm{{jv}}}_{\kappa}(\policya=\qdotdes, s_t=\qdot) = \kv (\qdotdes - \qdot) 
\end{equation}
\noindent
where we close the loop around $\qdot$, the current joint velocity (state), and $\kv$ is a vector of proportional gain (parameter $\kappa$).

\textbf{Joint Positions:}
For policies that output reference joint positions, $\policya = \qdes$, it is most straightforward to use a proportional-derivative (PD) controller that generates torques that increase with the joint position error and decrease with the current joint velocity. We also remove the dynamic effects of the mechanism by scaling the torques with the inertia matrix, $\massmatrix$~\cite{khatib1987unified}. The function to transform reference joint positions to joint torques is thus ($\textrm{jp} = \textrm{\it joint positions}$):

\begin{equation}
\taurobot = f^{\textrm{{jp}}}_{\kappa}(\qdes, \q, \qdot) = \massmatrix\left[\kp\qdelta - \kv\qdot\right]
\end{equation}
\noindent
where $\qdelta = \qdes - \q$ is the difference between current and desired joint configurations, which can be used as an alternative policy action space. $\kp$ and $\kv$ are vectors of proportional and derivative gains (parameters $\kappa$).

\textbf{End-Effector Pose:}
In the cases where the policy outputs the desired 6-D pose $\in \mathbb{SE}(3)$ of the robot in end-effector space, $\cartposd$, we can use an impedance-based PD controller to first derive an end-effector space acceleration to move towards the goal. To do that, $\cartposd$ can be decomposed into desired position, $\posdes \in \mathbb{R}^3$, and desired orientation, $\orides \in \mathbb{SO}(3)$. In the impedance-based PD controller, the end-effector acceleration increases with the difference between desired end-effector pose and current pose, $\pos$ and $\ori$, and decreases with the current end-effector velocity, $\cartvel$ and $\orivel$.

We then compute the robot actuations (joint torques) to achieve the desired end-effector space accelerations leveraging the kinematic and dynamic models of the robot with the dynamically-consistent operational space formulation~\cite{Khatib1995a}. First, we compute the wrenches at the end-effector that correspond to the desired accelerations, ${f}\in\mathbb{R}^{6}$.
Then, we map the wrenches in end-effector space ${f}$ to joint torque commands with the end-effector Jacobian at the current joint configuration ${J}={J}(\q)$: $\taurobot = {J}^T{f}$. 

Thus, the function that maps end-effector space position and orientation to low level robot commands is ($\textrm{ee} = \textrm{\it end-effector space}$):
\begin{equation}
\begin{aligned}
\taurobot &= f^{\textrm{{ee}}}_{\kappa}(\posdes, \orides, \pos, \ori, \cartvel, \orivel) \\
&= \jacpos[\lambdapos[\kppos(\posdes - \pos) - \kvpos\cartvel]] + \\
&~~~\jacori[\lambdaori\left[\kpori(\orides \ominus \ori) - \kvori\orivel\right]]
\end{aligned}
\end{equation}
\noindent
where $\lambdapos$ and $\lambdaori$ are the parts corresponding to position and orientation in $\Lambda \in \mathbb{R}^{6\times6}$, the inertial matrix in the end-effector frame that decouples the end-effector motions, ${J}_{\underscoredw{pos}}$ and ${J}_{\underscoredw{ori}}$ are the position and orientation parts of the end-effector Jacobian, and $\ominus$ corresponds to the subtraction in $\mathbb{SO}(3)$. The difference between current and desired position ($\cartdelta= \posdes - \pos$) and between current and desired orientation ($\oridelta = \orides \ominus \ori$) can be used as alternative policy action space, $\mathcal{A}$. $\kppos$, $\kvpos$, $\kpori$, and $\kvori$ are vectors of proportional and derivative gains for position and orientation (parameters $\kappa$), respectively.


\textbf{Variable Impedance End-Effector Space (VICES):}

Thus far, we have defined transformations between policy actions $\policya$ and robot commands $\taurobot$ are parameterized with $\kappa$. In these cases, parameters $\kappa$ are manually specified.
We observe that it is beneficial to augment the action space with these parameters to give the agent full control of the behavior. As discussed in Sec.~\ref{sec:tw}, this idea has been previously explored in joint space for $\kp$ and $\kv$~\cite{buchli2011learning}. In this paper, we propose to also turn the parameters of the end-effector space function ($\kppos$, $\kvpos$, $\kpori$, and $\kvori$) into policy outputs. We term this action space as \textit{Variable Impedance End-Effector Space}~(VICES). It enables the policy to learn both to predict the end-effector pose as a trajectory reference as well as to dynamically adapt the impedance gains along each of the six axes (rotation and translation) according to the phase of the task. 



\section{Experiments}
\label{sec:exp}
We conduct experiments in three application domains: a) free space Path Following~\cite{buchli2011learning}, b) manipulation of articulated mechanisms~\cite{kim2010impedance} and c) surface wiping~\cite{ott2008cartesian,Leidner2016RoboticAR}. These tasks are not only relevant applications in robotics, but also span different levels of task constraints from free motion to highly constrained contact-rich manipulation, which allows us to evaluate and compare the characteristics of the different action spaces for policy learning.

For these three tasks and for each of the evaluated action spaces we aim to answer the following questions: is the action space suitable for model-free RL? Is the learned policy physically efficient? 
Does the policy learned with a simulated robot transfer to a different simulated robot? Does a policy learned in simulation transfer to a real robot? To answer these questions we will use the following metrics and tests:
\begin{enumerate}[
    topsep=0pt,
    noitemsep,
    partopsep=0.5ex,
    parsep=0.5ex,
    leftmargin=*,
    itemindent=2.5ex
    ]
    \item \textbf{Sample efficiency and task completion}: samples required for the policy to succeed in the task and/or converge
    \item \textbf{Physical efficiency}: energy consumed by the robot when using the trained policy. We assume a proportional relationship between joint torques and electric power
    \item \textbf{Physical effort}: wrenches applied to the environment by the trained policy during contact-rich manipulation tasks
    \item \textbf{Transferability between robots}: does a robot achieve the task using a policy trained on a different robot?
    \item \textbf{Sim-to-real transfer} of contact-rich policies: does a real robot achieve the task using a policy trained in simulation?
\end{enumerate}

Our control framework is outlined in Fig~\ref{fig:controller}. In all experiments our policies output actions ($a \in \mathcal{A}$) at \SI{20}{\hertz}, while we send joint torque commands ($\taurobot \in \mathcal{U}=\mathcal{T}$) to the robot at \SI{500}{\hertz}. To generate torque commands at a higher frequency, the controllers use the constant desired goal from the policy while updating the current state of the robot, $s_{\textrm{\it robot}}$. In order to ensure smooth robot commands and generated motions, in all of our controllers we interpolate linearly between policy commands at consecutive time steps.

\subsection{Free-Space Motion - Path Following}
\label{ss:fst}

\noindent \textbf{Setup}. In this experiment we aim to measure the properties of different action spaces for tasks that do not involve any contact with the environment. 
Agent's goal is to follow a trajectory in free-space passing through four via-points. The via-points are placed on a virtual plane in front of the agent at a constant distance along the \emph{x} axis. The order and location of the via-points are fixed. We measure success as
the fraction of the four via-points that are passed through. 

This setup is a more complex version of the one via-point trajectory of Buchli~\etal~\cite{buchli2011learning}. This task can be solved kinematically without impedance control. However, we found that controlling the compliance of the robot could still offer benefits in this setup. 

\vspace{2pt}
\noindent \textbf{Reward Model}. This task is trained in two phases
: a first phase of task completion and a second phase of energy optimization. In the first phase, the agent is rewarded only to complete the task: to pass through the four via-points. In the second phase, the trained models from previous phase are further trained with the additional objective of optimizing their motion to reduce energy consumption.

In the first phase of the experiment, the agent is rewarded when it hits a via-point (it gets closer than $d_{\textrm{\it th}} = \SI{5}{\centi\meter}$). To help guide exploration, we also provide a small dense reward inversely proportional to the distance to the next via-point in the trajectory. 
Since the episodes continue after the task is completed, a task-completion bonus proportional to the remaining time steps was introduced to discourage the robot from unnecessarily extending the duration of the task. We train policies with this reward using the different action spaces to evaluate if they can learn to follow the free-space trajectory.

In the second phase of the experiment
, we explore if the action spaces can optimize for the additional objective of minimizing energy consumption without decreasing the quality of the first objective (passing through the via-points).
We include an energy consumption penalty to the previously defined reward function. To evaluate energy consumption we assume that the torques from the motors are proportional to electric current and the voltage is constant, and thus the amount of electric power scales proportional to the torque and the energy is its time integral.


\vspace{2pt}
\noindent \textbf{Observations}. We use as observations the pose and velocity of the end-effector in the robot reference frame, as well as the location of the via-points (and whether each one has been checked).

\vspace{2pt}
\noindent {\bf Evaluation}. We first evaluate each of the different action spaces in simulation, using a simulated Panda robot agent with five different random seeds. In the first phase of the experiment, we measure sample efficiency (reward as function of the iterations) and level of completion of the task (number of via-points crossed). In the second phase of the experiment, we also measure the total energy consumption and task success. In both phases we evaluate how the original trained policies transfer between robots.

\begin{figure}[t]
\centering
\includegraphics[width=0.9\linewidth]{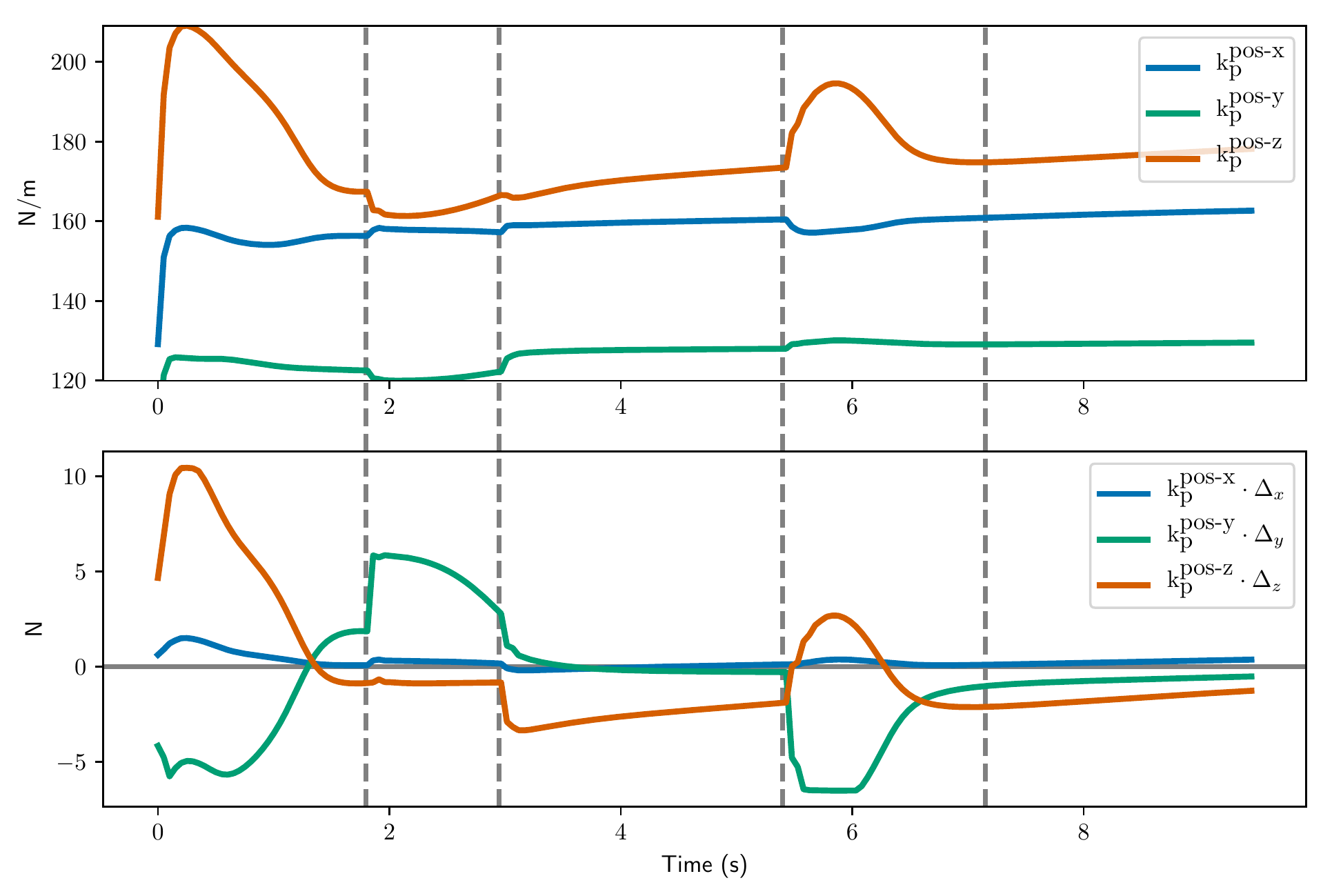}
\caption{Free space Path Following: Time evolution for a single episode of a policy using variable impedance in end-effector space; Impedance (stiffness, $k_p$) changes as via-points are checked as indicated by the vertical dotted lines; The four via-points of the trajectory are aligned with the same \emph{x} coordinate and thus the policy learns to not exert force in that dimension (blue curves); After checking a via-point, the policy increases stiffness ($k_p$) and combines it with the right desired displacement ($k_p\Delta x$) to generate motion in the direction to reach the next via-point of the trajectory}
\label{fig:fst}
\end{figure}

\begin{figure*}[ht]
\begin{subfigure}[t]{0.32\textwidth}
\includegraphics[width=0.95\linewidth]{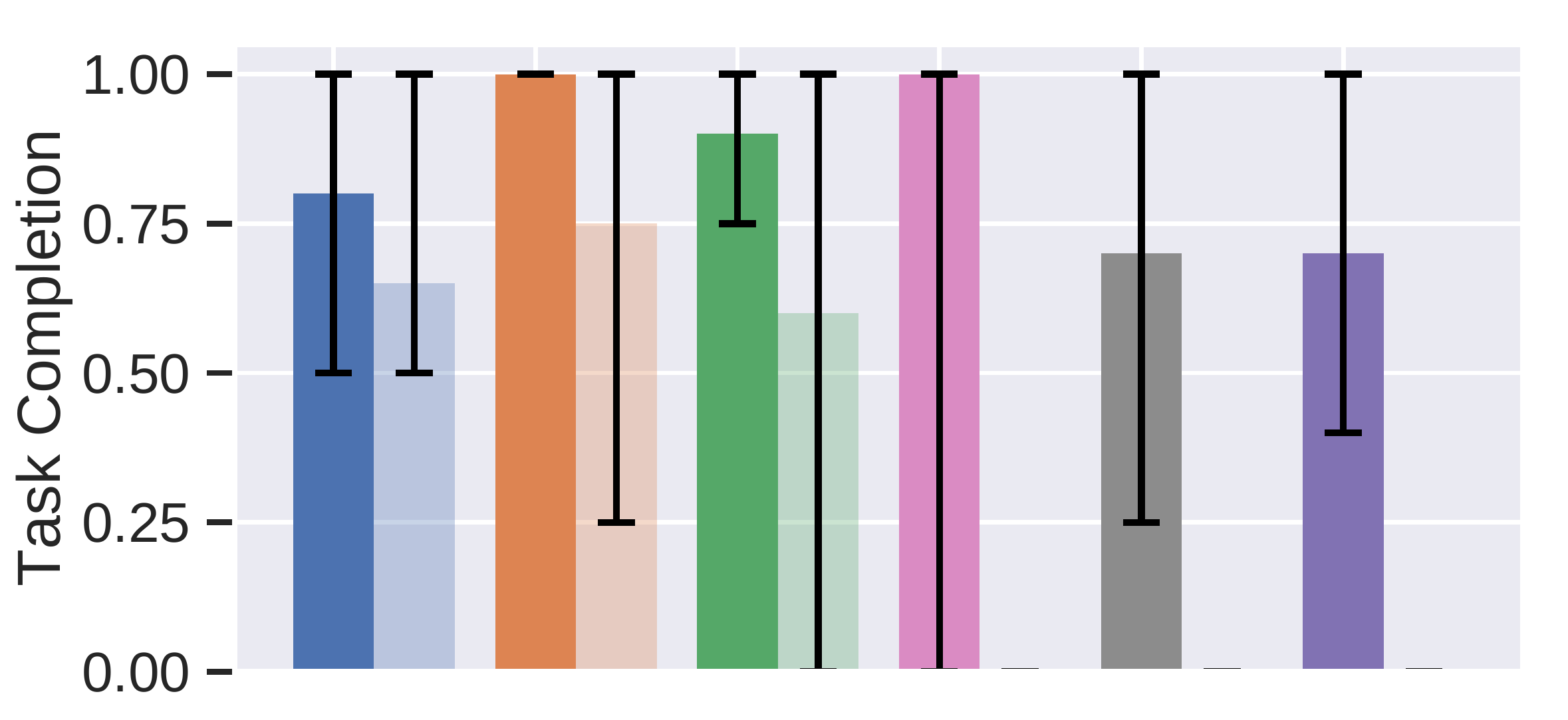}%
\caption{Path Following}
\label{fig:transf_traj}
\end{subfigure}
\hfill
\begin{subfigure}[t]{0.32\textwidth}
\includegraphics[width=0.95\linewidth]{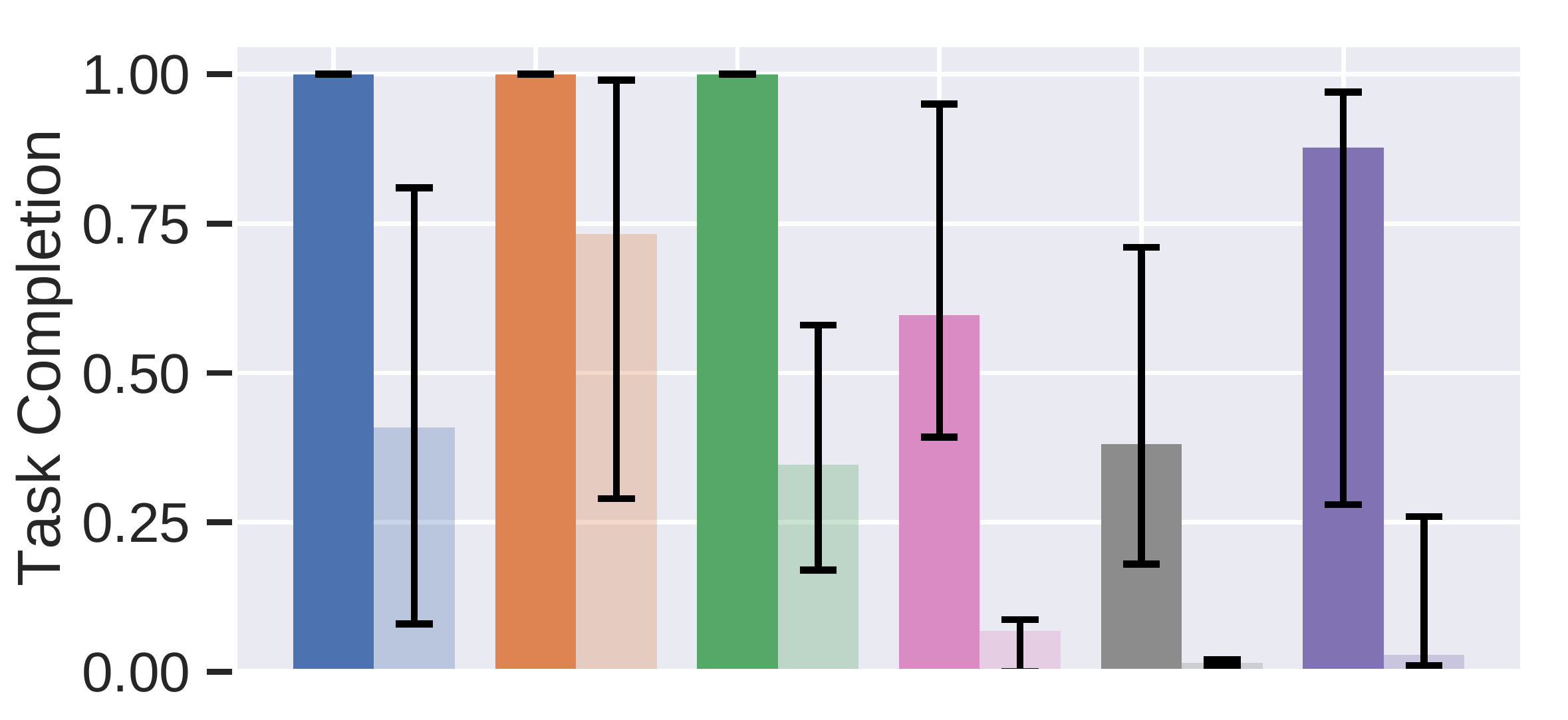}%
\caption{\small{Door Opening}}
\label{fig:transf_door}
\end{subfigure}
\hfill
\begin{subfigure}[t]{0.32\textwidth}
\includegraphics[width=0.95\linewidth]{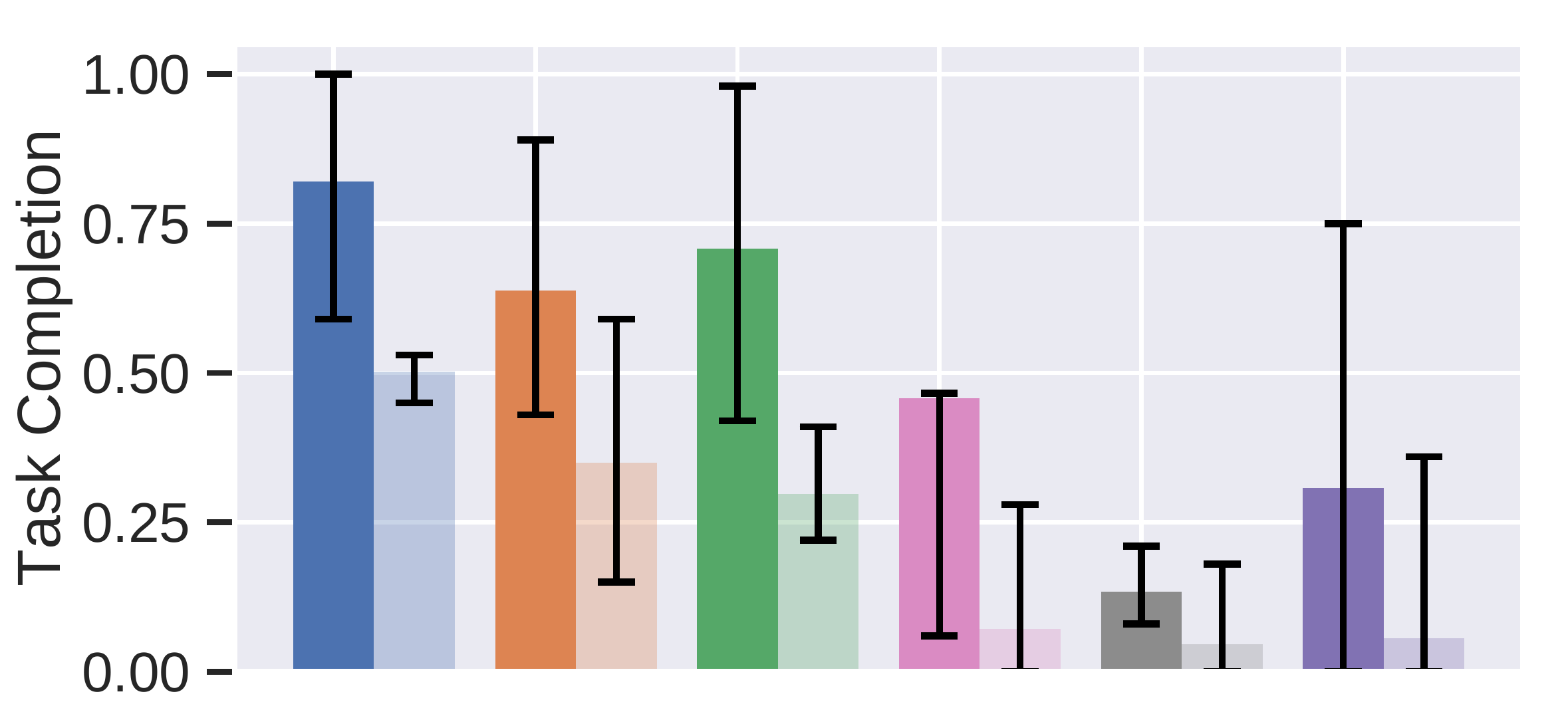}%
\caption{Surface Wiping}
\label{fig:transf_wipe}
\vspace{-10pt}
\end{subfigure}
\hfill
\begin{subfigure}[t]{0.99\textwidth}
\vspace{-2pt}
\includegraphics[trim={-4cm 5.5cm 4cm 0.5cm},clip,width=\linewidth]{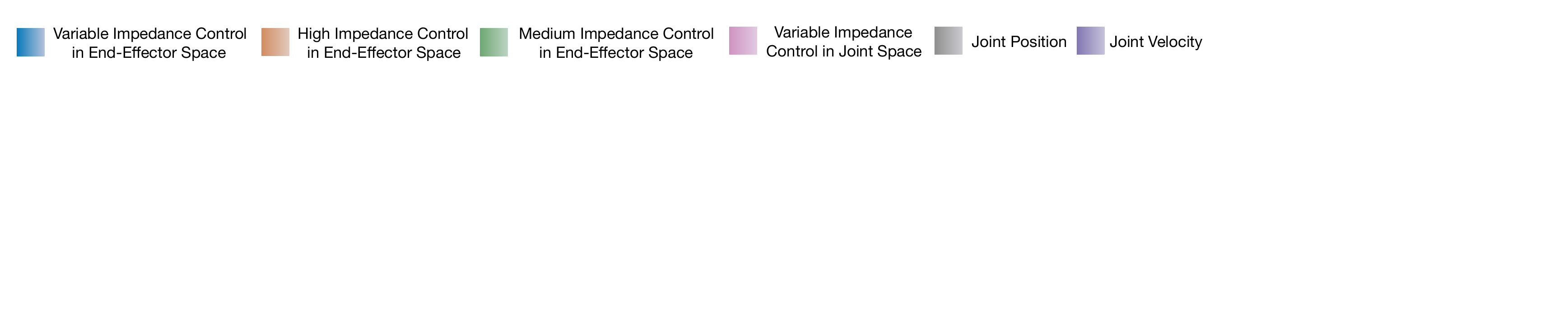}%
\end{subfigure}
\caption{Evaluation of the the task performance and transfer of policies learned on a robotic platform (Panda, dark color bars) to a different platform (Sawyer, lighter color bars) for a) Path Following (free space), b) door opening (kinematic constraints), and c) surface wiping (contact rich) tasks; Transfer is between simulated robots without retraining; The error bars indicate the range of percent task performance across the 5 seeds; Policies trained in the joint torque action space and in the end-effector low impedance action space did not learn to achieve the task in the original embodiment (Panda) and are not part of the transfer evaluation; Policies in joint space are unable to transfer between embodiments while policies in end-effector space transfer better because they are not dependant of the robot dynamics; Policy transfer is harder for tasks with increased contact with the environment, as the agent is more likely to hit joint limits}
\label{fig:transf}
\end{figure*}

\vspace{2pt}
\noindent \textit{Sample Efficiency and Task Completion}. Fig.~\ref{fig:training} (\emph{a}) shows the training curves for policies in each of the action spaces. All policies except the ones that output reference joint torques and end-effector poses resolved with a fixed low impedance controller were able to achieve the goal of the task: checking all 4 via-points (see Fig.~\ref{fig:transf_traj}). For the policies that achieve the task, the differences in reward value after convergence is simply a consequence of the termination bonus: some action spaces (e.g. desired end-effector poses resolved with high fixed impedance) allow for faster motion and thus faster completion of the task

We gain insights on how an RL policy exploits VICES for this task by observing the stiffness and damping over the course of an episode. Fig.~\ref{fig:fst} depicts the commands (the desired stiffness and the product of desired stiffness and delta position) from the policy trained with VICES for one episode after the first stage of training (before applying the energy penalty). The policy exploits the impedance (stiffness) to reach each via-point in the different portions of the trajectory. As it checks each via-point (indicated by the vertical bars in the figure), the impedance changes in the appropriate dimension to move quickly to the next via-point with enough stiffness to avoid overshooting.

\vspace{2pt}
\noindent \textit{Physical Efficiency}. We evaluate the physical efficiency of policies in different action spaces by comparing the total energy consumption of the agents at the end of the first phase and of the second phase of our experiment, where we add the energy penalty. We found that the policies using variable impedance in end-effector space as action space were the only end-effector space policies that consistently improved energy efficiency while maintaining task performance. Both the medium and high fixed impedance models became unstable, since the action space does not have sufficient degrees of freedom to optimize the motion to reduce energy consumption while still achieving the trajectory task. Note that since the low impedance model never achieved the task, it was not evaluated with energy penalties.

In joint space, the policies outputting actions in variable impedance space were also able to reduce the energy consumption significantly more than the controllers with fixed impedance, as expected~\cite{buchli2011learning}. They also reduced more energy than policies outputting variable impedance in end-effector space. This reflects that the policies outputting joint space commands resulting from the first phase of our experiment solved the task much faster (with higher energy consumption) than their end-effector counterparts and therefore had much more room for improvement when optimizing for energy efficiency. Therefore, the difference in absolute energy optimization between policies in joint and in end-effector space is an artifact of the difference in magnitude between end-effector space delta position limits and joint space delta angle limits (i.e., the joint space agents were originally allowed to move more at each time step).

\vspace{2pt}
\noindent \textit{Transferability}. We also evaluate how policies using different action spaces transfer in simulation from one robot to another through zero-shot transfer from the Panda robot to the Sawyer robot. The results are depicted in Fig.~\ref{fig:transf_traj}. As expected, we observed that after convergence only the policies using fixed and variable impedance in end-effector space could transfer directly between robots. The joint-space policies were not able to transfer due to the very different kinematics and dynamics of the two robot platforms. By using end-effector space control, we factor out the effects of the embodiment from the policy learning problem.

\subsection{Manipulation of Constrained Mechanisms - Door Opening}
\label{ss:mcm}

\noindent \textbf{Setup}. In this task, the robot has to learn how to manipulate a one DoF constrained mechanism, a door, to a specific configuration. The agent is equipped with a two-finger gripper it can use to hold the door handle. The door handle is a bar attached vertically on the door leaf. The gripper is closed, leaving a space between the fingers to cage the door handle while still allowing for rotation between handle and gripper. 

We ensure that the agent learns to interact in a controlled and safe manner. Hence instead of maximally opening the door, we set the goal to manipulate the door into a desired joint configuration ($\theta_{goal}$ = \SI{60}{\degree}). We measure success as the fraction of the total desired joint state achieved by the robot: $1 - \frac{|\theta_{door} - \theta_{goal}|}{\theta_{goal}}$. 




\vspace{2pt}
\noindent \textbf{Reward Model}.
We reward the agent when the door joint gets closer to the desired configuration. We provide additional constant reward if the configuration of the door is very close to the desired value (less than \SI{5}{\degree}). We penalize forces and torques exerted on the environment that go beyond the physical payload of the robot (\SI{40}{N}). We also penalize the agent for colliding with the environment with links other than the gripper and for going beyond its joint limits. For safety, the episode terminates when joint limits are violated. 

\vspace{2pt}
\noindent \textbf{Observations}.
We use as observation the pose and velocity of the robot's end-effector in the robot reference frame, as well as the door's angle and angular velocity. 

\vspace{2pt}
\noindent {\bf Evaluation:} We evaluate the different action spaces in simulation. We train an agent with a Panda robot embodiment for each action space with five different random seeds.

\vspace{2pt}
\noindent \textit{Sample Efficiency and Task Completion}. We first evaluate the different action spaces on their sample efficiency of learning the door-manipulation. The training results are depicted in Fig.~\ref{fig:training}, middle. The task success results for the door task can be found in Fig.~\ref{fig:transf_door}. 

We observe that policies that output end-effector space actions (with medium, variable, and high impedance) outperform policies in all other action spaces, in terms of achieving close to 100\% task success rate and higher rewards. In end-effector space, the policy resolved with an impedance controller with fixed medium stiffness and damping is able to learn the task at a faster rate than the variable impedance controller, as it is initialized with a suitable impedance to operate the door with the defined friction. However, policies outputting actions in the both aforementioned spaces reach similar rewards and task success rates at the end of training, as the policies that can vary impedance end up learning a suitable impedance for the task. 

While the policies resolved with an impedance controller with fixed high impedance parameters also achieve on average 100\% task success rate, their rewards are lower because they exert higher forces in the environment that is penalized. The policies resolved with an impedance controller with fixed low impedance parameters are not able to learn the door opening task because they cannot exert high enough forces to overcome the friction of the door and move it. The policies outputting joint velocity actions can reach up to  75\% task success, but the rewards are much lower than policies in VICES, as they often reach joint limits while opening the door. Policies outputting other joint-space actions (torques, positions) are unable to learn to exert enough force to open the door without reaching joint limits. 

\vspace{2pt}
\noindent \textit{Transferability}. We also evaluate the ability of policies in different action spaces to transfer from the Panda robot to the Sawyer robot. The results are shown in Fig.~\ref{fig:transf_door}, in lighter colors. Transferring policies for the door opening task is more complex than for the free-space Path Following task because the different robots' kinematics lead to very different task-space limitations, as well as very different joint limit constraints. Similar to results in the other tasks, policies trained in joint space are unable to transfer, since the kinematics of the robots differ substantially. The end-effector space policies are able to transfer much more successfully, as the end-effector space policies are able to abstract away the dynamics and kinematics of each specific robot model. There is still a performance drop, as the policies in end-effector space do not learn to account for the robots' different kinematic constraints (i.e. joint limits). 
\subsection{Contact-Rich Manipulation - Surface Wiping}
\label{ss:wiping}


\noindent \textbf{Setup}. In this experiment the goal is to wipe a table whose surface location is unknown. The agents are equipped with a wiping tool, resembling a scrubber or a whiteboard eraser (see Fig.~\ref{fig1}). In the simulator, the tool is modeled as a soft material that creates contact forces that increase proportionally to the penetration into the tool's surface. The material to wipe is modeled as a set of small elements of a color different from the table. The elements are placed randomly on the table surface to form a continuous ``stain'' and are marked initially as \emph{unwiped}. They become \emph{wiped} if the wiping tool passes through them, which also causes them to disappear visually. Note that since the elements are modeled as very thin (\SI{1}{\milli\meter} height) cylinders resting on the table's surface, the agent needs to press the tool against the surface so as to be able to wipe elements. Success rate is measured as the fraction of the elements wiped. The sliding coefficient of friction of the table acting along both axes of the tangent plane is sampled uniformly between $1.0$ and $0.1$. The initial location of the agent above the table is randomized. 

\vspace{2pt}
\noindent \textbf{Reward Model}. The main reward comes from wiping off elements. We also provide additional reward for wiping off all the elements. Additionally, to help during the initial phases of exploration, we give the agent a small reward for maintaining contact with the table. Finally, since we aim to generate safe solutions that can directly be tested on the real robot, we slightly penalize the agent for applying forces over the payload of the real robot (\SI{40}{\newton}), and harshly penalize the agent for reaching joint limits or colliding with the table with parts other than the wiping tool. If such collisions occur, the episode ends and the tasks restarts.

\vspace{2pt}
\noindent \textbf{Observations}. There is no straightforward way to represent the state of a wiping task. Instead, we directly use visual observations: $48\times48\times3 $ RGB images of the wiping scene generated in our simulator, and obtained from a camera on our real robot platform for the simulation-to-real transfer experiments. As in previous experiments, we also provide the pose and velocity of the end-effector.

\vspace{2pt}
\noindent {\bf Evaluation:} We first evaluate the different action spaces on simulation. We train an agent with a Panda robot embodiment for each action space with five different random seeds.

\vspace{2pt}
\noindent \textit{Sample Efficiency and Task Completion}. Fig.~\ref{fig:training}, right, shows the convergence of the agents with different action spaces. We observe that the agents with variable impedance in end-effector space converge faster and achieve higher reward. The higher reward is obtained thanks to a lower penalty for applying excessive force on the table since the agents can learn to appropriately adapt their stiffness. The mean force applied by the policies with variable impedance in end-effector space is \SI{28}{\newton}, less than the robot's payload. Agents using other action spaces apply higher mean force or not enough to wipe the table. In terms of task completion, the results are depicted in Fig.~\ref{fig:transf_wipe}, dark colors. Policies outputting actions in VICES achieve the highest ratio of wiped units.

\begin{figure}[t!]
\includegraphics[width=0.42\textwidth]{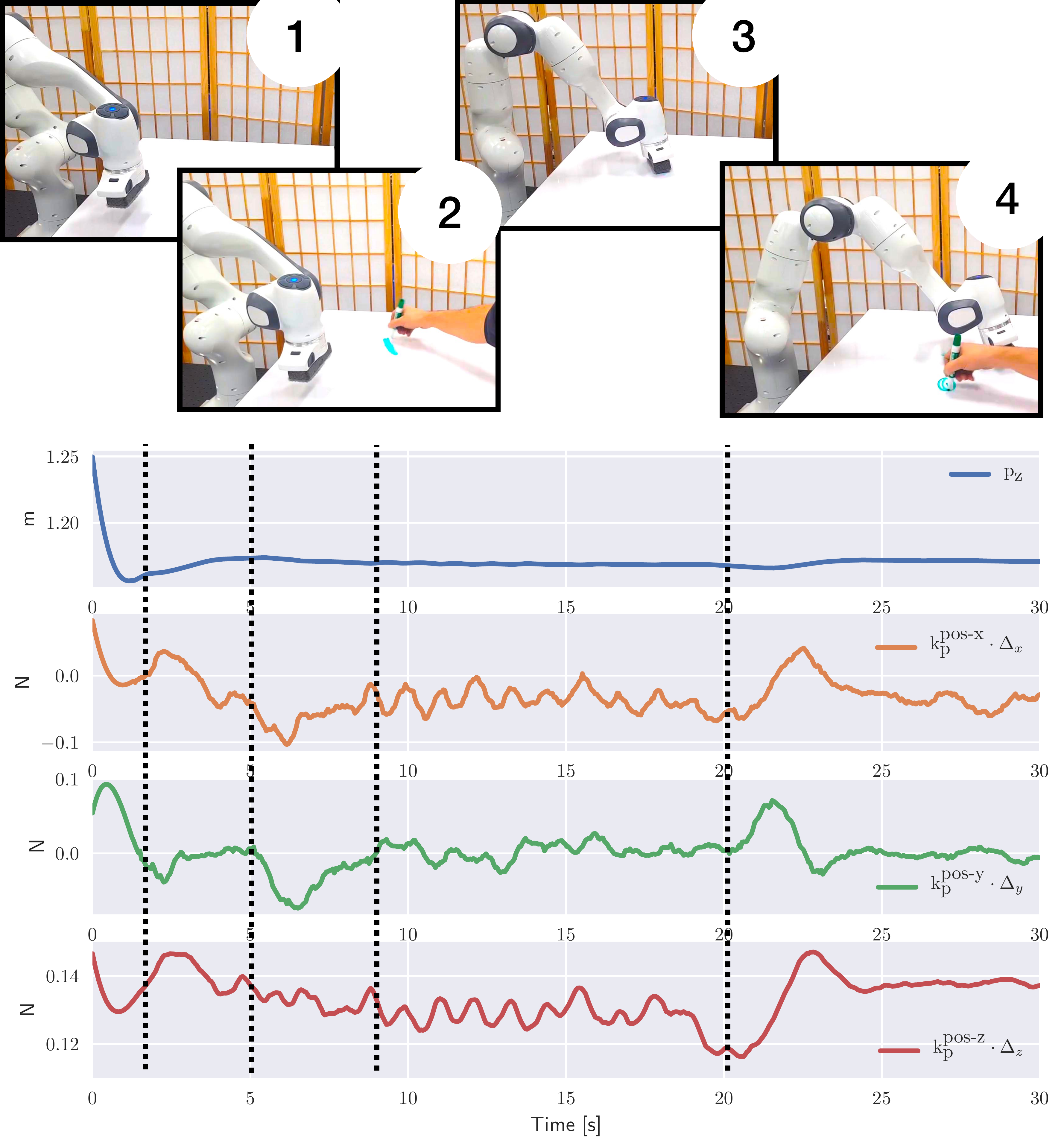}
\caption{Sim2Real transfer: we apply directly the policy learned in simulation with action in variable impedance in end-effector space to the real robot; pictures correspond to dotted vertical lines in the plots; 1) the robot pushes towards the table in a compliant manner (height over the table in the top plot, force in $z$, bottom plot); the experimenter marks different parts of the whiteboard, 2) and 4), and the robot reacts 3) moving towards the area following the forces in $x$ and $y$ (second and third rows)}
\label{fig:s2r}
\end{figure}

\vspace{2pt}
\noindent \textit{Transferability}. We also evaluate if the policies learned with the Panda robot embodiment transfer directly to the Sawyer robot in simulation. Fig.~\ref{fig:transf_wipe}, depict the results of the policy transfer between robots. Policies trained in variable impedance in end-effector space transfer better than policies in any other space since the policy is independent of the robot embodiment. However, there is a significant drop in performance due to the different forces generated by the different embodiments.

\vspace{2pt}
\noindent \textit{Simulation-to-real transfer}. 
In a final experiment we evaluate if the policies trained in simulation can be used on the real robot without any retraining. We use in our experiment the best performing simulation policy. The goal in the real world is to wipe a whiteboard painted with a marker. Since our focus is on the evaluation of the action space and not on learning a representation of the image, we convert the real images into fake simulated images by superimposing the results of a color segmentation for the colored parts of the table on an image from the simulator where the robot configuration is set to track the real robot. As a safety precaution, we stop the robot if the payload is exceeded. We note that the robot does not use any direct force sensing during the experiments. 

We initialize the robot to the same location and run ten trials each with a different part of the whiteboard painted. One example of the run can be seen in Fig.~\ref{fig:s2r} and more runs in the video attachment. We assume a successful trial when the robot wipes more than 3/4 of the painted line. The robot wipes successfully the board in 8 of the 10 trials. In one of the failed trials the robot moved abruptly and triggered the safety mechanism. In another trial the robot did not wipe the mark entirely. These results indicate that the policies trained with VICES can transfer seamlessly to real world by exploiting the knowledge of the dynamics model of the robot.


\section{Conclusion} 
\label{sec:conclusion}
Reinforcement Learning (RL) as a family of algorithms has ushered in impressive results in generalization, yet principled evaluation on how to choose action spaces to learn control policies is missing. We presented a thorough evaluation of the effect of the choice of action space on learning policies in RL for tasks without contact, with kinematic constraints and contact-rich manipulation tasks. 
We also presented variable impedance in end-effector space (VICES) as an efficient choice of action space for RL and showed empirically that, even when contact conditions are dynamically variable during the task, this model outperforms other action space choices on sample efficiency, energy consumption, and safety. We also showed that, thanks to the subtraction of the dynamic effects of the embodiment, using variable impedance in end-effector space we can transfer policies learned in simulation to other simulated robots and to a real robot without fine tuning.



\renewcommand*{\bibfont}{\footnotesize}

\begin{flushright}
\printbibliography 

@article{Khatib1995a,
author = {Khatib, O.},
journal = {IJRR},
number = {1},
pages = {19--36},
title = {{Inertial Properties in Robotic Manipulation: An Object-Level Framework}},
volume = {14},
year = {1995}
}

@inproceedings{gu2017deep,
  title={Deep reinforcement learning for robotic manipulation with asynchronous off-policy updates},
  author={Gu, Shixiang and Holly, Ethan and Lillicrap, Timothy and Levine, Sergey},
  booktitle={ICRA},
  pages={3389--3396},
  year={2017},
  organization={IEEE}
}

@inproceedings{zhu2018reinforcement,
  title={Reinforcement and imitation learning for diverse visuomotor skills},
  author={Zhu, Yuke and Wang, Ziyu and Merel, Josh and Rusu, Andrei and Erez, Tom and Cabi, Serkan and Tunyasuvunakool, Saran and Kram{\'a}r, J{\'a}nos and Hadsell, Raia and de Freitas, Nando and others},
  booktitle={RSS},
  year={2018}
}

@article{buchli2011learning,
  title={Learning variable impedance control},
  author={Buchli, Jonas and Stulp, Freek and Theodorou, Evangelos and Schaal, Stefan},
  journal={IJRR},
  volume={30},
  number={7},
  pages={820--833},
  year={2011},
  publisher={SAGE Publications Sage UK: London, England}
}

@INPROCEEDINGS{6907861, 
author={M. {Li} and H. {Yin} and K. {Tahara} and A. {Billard}}, 
booktitle={ICRA}, 
title={Learning object-level impedance control for robust grasping and dexterous manipulation}, 
year={2014}, 
volume={}, 
number={}, 
pages={6784-6791}, 
month={5},}

@inproceedings{lee2015learning,
  title={Learning force-based manipulation of deformable objects from multiple demonstrations},
  author={Lee, Alex X and Lu, Henry and Gupta, Abhishek and Levine, Sergey and Abbeel, Pieter},
  booktitle={ICRA},
  pages={177--184},
  year={2015},
  organization={IEEE}
}

@inproceedings{Leidner2016RoboticAR,
  title={Robotic Agents Representing, Reasoning, and Executing Wiping Tasks for Daily Household Chores},
  author={Daniel Leidner and Wissam Bejjani and Alin Albu-Sch{\"a}ffer and Michael Beetz},
  booktitle={AAMAS},
  year={2016}
}

@article{kim2010impedance,
  title={Impedance learning for robotic contact tasks using natural actor-critic algorithm},
  author={Kim, Byungchan and Park, Jooyoung and Park, Shinsuk and Kang, Sungchul},
  journal={Transactions on Systems, Man, and Cybernetics},
  volume={40},
  number={2},
  pages={433--443},
  year={2010},
  publisher={IEEE}
}

@article{Rey2018,
author = {Rey, Joel and Kronander, Klas and Farshidian, Farbod and Buchli, Jonas and Billard, {\textperiodcentered} Aude},
title = {{Learning motions from demonstrations and rewards with time-invariant dynamical systems based policies}},
volume = {42},
year = {2018}
}

@INPROCEEDINGS{5648931, 
author={S. {Calinon} and I. {Sardellitti} and D. G. {Caldwell}}, 
booktitle={IROS}, 
title={Learning-based control strategy for safe human-robot interaction exploiting task and robot redundancies}, 
year={2010}, 
volume={}, 
number={}, 
pages={249-254}, 
month={10},}

@ARTICLE{6636303, 
author={K. {Kronander} and A. {Billard}}, 
journal={Transactions on Haptics}, 
title={Learning Compliant Manipulation through Kinesthetic and Tactile Human-Robot Interaction}, 
year={2014}, 
volume={7}, 
number={3}, 
pages={367-380}, 
month={7},}

@article{Mitrovic2011,
author = {Mitrovic, Djordje and Klanke, Stefan and Vijayakumar, Sethu},
journal = {IJRR},
number = {5},
pages = {556--573},
title = {{Learning impedance control of antagonistic systems based on stochastic optimization principles}},
volume = {30},
year = {2011}
}

@article{Viereck2018,
author = {Viereck, Julian and Kozolinsky, Jules and Herzog, Alexander and Righetti, Ludovic},
journal = {RAL},
title = {{Learning a Structured Neural Network Policy for a Hopping Task}},
year = {2018}
}

@article{Abu-Dakka2018,
author = {Abu-Dakka, Fares J. and Rozo, Leonel and Caldwell, Darwin G.},
journal = {Robotics and Autonomous Systems},
month = {11},
pages = {156--167},
title = {{Force-based variable impedance learning for robotic manipulation}},
volume = {109},
year = {2018}
}

@article{rozo2016learning,
  title={Learning physical collaborative robot behaviors from human demonstrations},
  author={Rozo, Leonel and Calinon, Sylvain and Caldwell, Darwin G and Jimenez, Pablo and Torras, Carme},
  journal={Transactions on Robotics},
  volume={32},
  number={3},
  year={2016},
  publisher={IEEE}
}

@article{ruckert2013learned,
  title={Learned graphical models for probabilistic planning provide a new class of movement primitives},
  author={R{\"u}ckert, Elmar A and Neumann, Gerhard and Toussaint, Marc and Maass, Wolfgang},
  journal={Frontiers in computational neuroscience},
  volume={6},
  pages={97},
  year={2013},
  publisher={Frontiers}
}

@inproceedings{mnih2016asynchronous,
  title={Asynchronous methods for deep reinforcement learning},
  author={Mnih, Volodymyr and Badia, Adria Puigdomenech and Mirza, Mehdi and Graves, Alex and Lillicrap, Timothy and Harley, Tim and Silver, David and Kavukcuoglu, Koray},
  booktitle={ICML},
  pages={1928--1937},
  year={2016}
}

@article{schulman2017proximal,
  title={Proximal policy optimization algorithms},
  author={Schulman, John and Wolski, Filip and Dhariwal, Prafulla and Radford, Alec and Klimov, Oleg},
  journal={arXiv},
  year={2017}
}

@book{sutton2018reinforcement,
  title={Reinforcement learning: An introduction},
  author={Sutton, Richard S and Barto, Andrew G},
  year={2018},
  publisher={MIT press}
}

@inproceedings{schulman2015trust,
  title={Trust Region Policy Optimization.},
  author={Schulman, John and Levine, Sergey and Abbeel, Pieter and Jordan, Michael I and Moritz, Philipp},
  booktitle={Icml},
  volume={37},
  pages={1889--1897},
  year={2015}
}

@article{lillicrap2015continuous,
  title={Continuous control with deep reinforcement learning},
  author={Lillicrap, Timothy P and Hunt, Jonathan J and Pritzel, Alexander and Heess, Nicolas and Erez, Tom and Tassa, Yuval and Silver, David and Wierstra, Daan},
  journal={arXiv preprint arXiv:1509.02971},
  year={2015}
}

@article{part1985impedance,
  title={Impedance control: An approach to manipulation},
  author={Hogan, Neville},
  journal={Journal of dynamic systems, measurement, and control},
  volume={107},
  pages={17},
  year={1985}
}

@inproceedings{Peng:2017:LLS:3099564.3099567,
 author = {Peng, Xue Bin and van de Panne, Michiel},
 title = {Learning Locomotion Skills Using DeepRL: Does the Choice of Action Space Matter?},
 booktitle = {SIGGRAPH},
 year = {2017},
 location = {USA},
 publisher = {ACM},
}

@book{ott2008cartesian,
  title={Cartesian impedance control of redundant and flexible-joint robots},
  author={Ott, Christian},
  year={2008},
  publisher={Springer}
}

@article{khatib1987unified,
  title={A unified approach for motion and force control of robot manipulators: The operational space formulation},
  author={Khatib, Oussama},
  journal={IEEE Journal on Robotics and Automation},
  volume={3},
  number={1},
  pages={43--53},
  year={1987},
  publisher={IEEE}
}

@article{kroger2004compliant,
  title={Compliant motion programming: The task frame formalism revisited},
  author={Kr{\"o}ger, Torsten and Finkemeyer, Bernd and Thomas, Ulrike and Wahl, Friedrich M},
  journal={Mechatronics \& Robotics, Aachen, Germany},
  year={2004},
  publisher={Citeseer}
}

@ARTICLE{4308708, 
author={M. T. {Mason}}, 
journal={Transactions on Systems, Man, and Cybernetics}, 
title={Compliance and Force Control for Computer Controlled Manipulators}, 
year={1981}, 
volume={11}, 
number={6}, 
pages={418-432}, 
doi={10.1109/TSMC.1981.4308708}, 
ISSN={0018-9472}, 
month={6},}

@ARTICLE{508440, 
author={H. {Bruyninckx} and J. {De Schutter}}, 
journal={Transactions on Robotics and Automation}, 
title={Specification of force-controlled actions in the "task frame formalism"-a synthesis}, 
year={1996}, 
volume={12}, 
number={4}, 
pages={581-589}, 
doi={10.1109/70.508440}, 
ISSN={1042-296X}, 
month={8},}

@article{Li2018,
author = {Li, Yanan and Ganesh, Gowrishankar and Jarrasse, Nathanael and Haddadin, Sami and Albu-Schaeffer, Alin and Burdet, Etienne},
journal = {Transactions on Robotics},
number = {5},
pages = {1170--1182},
pmid = {18502882},
title = {{Force, Impedance, and Trajectory Learning for Contact Tooling and Haptic Identification}},
volume = {34},
year = {2018}
}

@article{levine2016end,
  title={End-to-end training of deep visuomotor policies},
  author={Levine, Sergey and Finn, Chelsea and Darrell, Trevor and Abbeel, Pieter},
  journal={JMLR},
  volume={17},
  number={1},
  year={2016},
}

@inproceedings{sen2016automating,
  title={Automating Multiple-Throw Multilateral Surgical Suturing with a Mechanical Needle Guide and Sequential Convex Optimization},
  author={Sen*, Siddarth and Garg*, Animesh and Gealy, David V and McKinley, Stephen and Jen, Yiming and Goldberg (* equal contribution), Ken},
  booktitle={ICRA},
  year={2016}
}

@article{schaal2003computational,
  title={Computational approaches to motor learning by imitation},
  author={Schaal, Stefan and Ijspeert, Auke and Billard, Aude},
  journal={Philosophical Transactions of the Royal Society of London. Series B: Biological Sciences},
  volume={358},
  number={1431},
  pages={537--547},
  year={2003},
  publisher={The Royal Society}
}

@article{khansari2016adaptive,
  title={Adaptive human-inspired compliant contact primitives to perform surface--surface contact under uncertainty},
  author={Khansari, Mohammad and Klingbeil, Ellen and Khatib, Oussama},
  journal={IJRR},
  volume={35},
  number={13},
  pages={1651--1675},
  year={2016},
  publisher={SAGE Publications Sage UK: London, England}
}

@inproceedings{kroemer2015towards,
  title={Towards learning hierarchical skills for multi-phase manipulation tasks},
  author={Kroemer, Oliver and Daniel, Christian and Neumann, Gerhard and Van Hoof, Herke and Peters, Jan},
  booktitle={ICRA},
  year={2015},
  organization={IEEE}
}

@inproceedings{lee2019making,
  title={Making Sense of Vision and Touch: Self-Supervised Learning of Multimodal Representations for Contact-Rich Tasks},
  author={Lee, Michelle A and Zhu, Yuke and Srinivasan, Krishnan and Shah, Parth and Savarese, Silvio and Fei-Fei, Li and Garg, Animesh and Bohg, Jeannette},
  booktitle={ICRA},
  year={2019}
}

@inproceedings{harrison2017adapt,
  title={ADAPT: Zero-Shot Adaptive Policy Transfer for Stochastic Dynamical Systems},
author={Harrison*, James and Garg*, Animesh and Ivanovic, Boris and Zhu, Yuke and Savarese, Silvio and Fei-Fei, Li and Pavone (* equal contribution), Marco},
  booktitle={ISRR},
  year={2017},
  organization={SPRINGER STAR}
}

@INPROCEEDINGS{Mrinal:2011,
author={M. Kalakrishnan and L. Righetti and P. Pastor and S. Schaal},
booktitle={IROS},
title={Learning force control policies for compliant manipulation},
year={2011},
doi={10.1109/IROS.2011.6095096},
ISSN={2153-0866}
}

@article{haarnoja2018soft,
  title={Soft Actor-Critic Algorithms and Applications},
  author={Haarnoja, Tuomas and Zhou, Aurick and Hartikainen, Kristian and Tucker, George and Ha, Sehoon and Tan, Jie and Kumar, Vikash and Zhu, Henry and Gupta, Abhishek and Abbeel, Pieter and others},
  journal={arXiv preprint arXiv:1812.05905},
  year={2018}
}

@article{vevcerik2017leveraging,
  title={Leveraging demonstrations for deep reinforcement learning on robotics problems with sparse rewards},
  author={Ve{\v{c}}er{\'\i}k, Matej and Hester, Todd and Scholz, Jonathan and Wang, Fumin and Pietquin, Olivier and Piot, Bilal and Heess, Nicolas and Roth{\"o}rl, Thomas and Lampe, Thomas and Riedmiller, Martin},
  journal={arXiv preprint arXiv:1707.08817},
  year={2017}
}

@article{2017_rss_system,
  title = {Real-time Perception meets Reactive Motion Generation},
  author = {Kappler, Daniel and Meier, Franziska and Issac, Jan and Mainprice, Jim and Garcia Cifuentes, Cristina and W{\"u}thrich, Manuel and Berenz, Vincent and Schaal, Stefan and Ratliff, Nathan and Bohg, Jeannette},
  journal = {IEEE Robotics and Automation Letters},
  volume = {3},
  number = {3},
  pages = {1864-1871},
  month = jul,
  year = {2018},
  url = {https://arxiv.org/abs/1703.03512},
  month_numeric = {7}
}

@InProceedings{18-toussaint-RSS,
title = {Differentiable Physics and Stable Modes for Tool-Use and
Manipulation Planning},
author  = {Marc Toussaint and Kelsey R Allen and Kevin A Smith and
Josh B Tenenbaum},
booktitle  = {RSS},
year = {2018},
}

@article{Ijspeert:2013:DMP,
 author = {Ijspeert, Auke Jan and Nakanishi, Jun and Hoffmann, Heiko and Pastor, Peter and Schaal, Stefan},
 title = {Dynamical Movement Primitives: Learning Attractor Models for Motor Behaviors},
 journal = {Neural Comput.},
 volume = {25},
 number = {2},
 year = {2013},
 numpages = {46},
 publisher = {MIT Press},
 address = {Cambridge, MA, USA},
}

@article{Ludo_Contact_2013,
author = {Ludovic Righetti and Jonas Buchli and Michael Mistry and Mrinal Kalakrishnan and Stefan Schaal},
title ={Optimal distribution of contact forces with inverse-dynamics control},
journal = {IJRR},
volume = {32},
number = {3},
pages = {280-298},
year = {2013}
}

@inproceedings{Calinon08IROS,
  author="S. Calinon and A. Billard",
  title="A Probabilistic Programming by Demonstration Framework Handling Skill Constraints in Joint Space and Task Space",
  booktitle="Proc. {IEEE/RSJ} Intl Conf. on Intelligent Robots and Systems ({IROS})",
  year="2008",
  month="9",
  location="Nice, France",
  pages="367--372"
}

@INPROCEEDINGS{DMP_Initial_2002,
author={A. J. {Ijspeert} and J. {Nakanishi} and S. {Schaal}},
booktitle={ICRA},
title={Movement imitation with nonlinear dynamical systems in humanoid robots},
year={2002},
volume={2},
number={},
pages={1398-1403 vol.2},
doi={10.1109/ROBOT.2002.1014739},
ISSN={},
month={5},}

@article{Argall:2009:SRL,
 author = {Argall, Brenna D. and Chernova, Sonia and Veloso, Manuela and Browning, Brett},
 title = {A Survey of Robot Learning from Demonstration},
 journal = {Robot. Auton. Syst.},
 issue_date = {May, 2009},
 volume = {57},
 number = {5},
 month = may,
 year = {2009},
 issn = {0921-8890},
 pages = {469--483},
 numpages = {15}
}

@article{theodorou2010generalized,
  title={A generalized path integral control approach to reinforcement learning},
  author={Theodorou, Evangelos and Buchli, Jonas and Schaal, Stefan},
  journal={JMLR},
  volume={11},
  number={Nov},
  pages={3137--3181},
  year={2010}
}

@inproceedings{thananjeyan2017multilateral,
  title={Multilateral surgical pattern cutting in 2d orthotropic gauze with deep reinforcement learning policies for tensioning},
  author={Thananjeyan, Brijen and Garg, Animesh and Krishnan, Sanjay and Chen, Carolyn and Miller, Lauren and Goldberg, Ken},
  booktitle={ICRA},
  pages={2371--2378},
  year={2017},
  organization={IEEE}
}
\end{flushright}

\end{document}